\documentclass[5p]{elsarticle}

\usepackage{lineno,hyperref}
\modulolinenumbers[5]

\makeatletter
\def\ps@pprintTitle{%
 \let\@oddhead\@empty
 \let\@evenhead\@empty
 \def\@oddfoot{}%
 \let\@evenfoot\@oddfoot}
\makeatother

\journal{Knowledge-Based Systems}

\usepackage{amsmath}
\usepackage{booktabs}
\usepackage{multirow}
\usepackage{stfloats}
\newtheorem{definition}{Definition}
\newtheorem{problem}{Problem}

\begin{document}

\begin{frontmatter}

  \title{DouFu: A \underline{Dou}ble \underline{Fu}sion Joint Learning Method For Driving Trajectory Representation}

  \author[mymainaddress,mysecondaryaddress]{Han Wang}

  \author[mymainaddress,mysecondaryaddress]{Zhou Huang\corref{mycorrespondingauthor}}
  \cortext[mycorrespondingauthor]{Corresponding author}
  \ead{huangzhou@pku.edu.cn}

  \author[mymainaddress,mysecondaryaddress]{Xiao Zhou}
  \author[mymainaddress,mysecondaryaddress]{Ganmin Yin}
  \author[mymainaddress,mysecondaryaddress]{Yi Bao}
  \author[mymainaddress,mysecondaryaddress]{Yi Zhang}

  \address[mymainaddress]{Institute of Remote Sensing and Geographical Information Systems, Peking University, Beijing, China}
  \address[mysecondaryaddress]{Beijing Key Lab of Spatial Information Integration \& Its Applications, Peking University, Beijing, China}

  \begin{abstract}
    Driving trajectory representation learning is of great significance for various location-based services, such as driving pattern mining and route recommendation. However, previous representation generation approaches tend to rarely address three challenges: 1) how to represent the intricate semantic intentions of mobility inexpensively; 2) complex and weak spatial-temporal dependencies due to the sparsity and heterogeneity of the trajectory data; 3) route selection preferences and their correlation to driving behavior. In this paper, we propose a novel multimodal fusion model, DouFu, for trajectory representation joint learning, which applies multimodal learning and attention fusion module to capture the internal characteristics of trajectories. We first design movement, route, and global features generated from the trajectory data and urban functional zones and then analyze them respectively with the attention encoder
    or feed forward network. The attention fusion module incorporates route features with movement features to create a better spatial-temporal embedding. With the global semantic feature, DouFu produces a comprehensive embedding for each trajectory. We evaluate representations generated by our method and other baseline models on classification and clustering tasks. Empirical results show that DouFu outperforms other models in most of the learning algorithms like the linear regression and the support vector machine by more than 10\%.
  \end{abstract}


\end{frontmatter}


\section{Introduction}\label{sec:introduction}
With GPS-devices being widely equipped and a large amount of location based applications, enormous quantities of driving data are being generated and captured in the form of trajectories. A trajectory is traditionally described as a sequence of spatially located points with time stamps. As the temporal record of interactions between users and spatial environment, driving trajectories are capable of demonstrating users' behaviour characteristic and traveling intention, which can be exploited further to a various kind of tasks like user portrait analysis\cite{wang2018you, cao2019habit2vec}, next location recommendation\cite{Liu2016PredictingTN,bao2021bilstm}, human activity classification\cite{zhou2018trajectory} etc. Patterns mined from trajectories can also offer instructions and advised for city construction planing. How to represent a trajectory inside machines has been a key issue before completing the data mining tasks. Besides spatial and temporal movement, the trajectory is semantically rich which can tell us what the driver is looking for. Despite its significance, few current representation models attempt to incorporate semantic information.


Recently, learning representation of trajectories in a fixed length latent space has been intensively studied. The various length trajectory data cannot share the same fixed length representation space, which makes it difficult to manage and evaluate the characteristic and relevant relationship. But in the learned feature space, each trajectory will be converted into a fixed length vector called embedding which represents this trajectory. A good representation learning model can maintain the relative relationships between trajectories, mapping them into a distance between embeddings in the feature space\cite{bengio2013representation}. Deep learning methods have been effectively utilized in many trajectory data mining tasks including representation learning\cite{dong2016characterizing, dong2017autoencoder,kieu2018distinguishing}. A variety of downstream tasks can be accomplished with trajectory embeddings produced by learning models. For trajectories, embeddings make it easy to measure similarities about them\cite{liu2010towards, li2018deep}. In the latent feature space, the euclidean distance between embeddings may reveal the relationship between them, which is useful for trajectory groups partitioning and clustering. Meanwhile, trajectory analysis could demonstrate users' characteristic and preference about traveling, which will benefit for location-based recommendation system\cite{wang2018you}.

However, current models mainly concentrate only on the spatial and temporal characteristics of trajectories, which lacks taking into account rich semantic information. As a spatial-temporal interaction sequence for drivers who intend to obtain the service in destination along a specific route, the trajectory should be able to fully depict the user's intention to travel. Besides the departure time and destination location, the functions provided by destination could also tell us what the drivers truly need. Additionally, route selection matters in the travel process. Different drivers may choose different routes. Some drivers may prefer a faster way, but others like a shorter one. Actually, only a few them would like to take the fastest route\cite{letchner2006trip}. The selection of a route demonstrates the characteristics of users, but it is rarely mentioned in existing research. Furthermore, several data inputs may not be perfectly independent of each other, especially those which occur in the same spatial-temporal environment. It is necessary to investigate the associations between various modalities in the learning model. A correlation model can exploit different attributes to calibrate each other for better performance.

Thus, it is possible to analyze semantic information of trajectories from such perspectives. We can convert a trajectory from these perspectives into different independent inputs and apply methods respectively to extract features from them. These modality features are capable of coordinating with each other to improve the performance of downstream tasks. In this paper, we aim to make full use of the multimodal features of a trajectory to produce a better representation.

To address these issues, we proposed a novel representation learning model, \textbf{DouFu}, for driving trajectories inspired by multimodal fusion and attention mechanism. The spatial-temporal and semantic information of the trajectory can be merged in the process of multimodal learning to acquire a good computational representation for driving in a latent feature space, providing a good foundation for downstream data mining and pattern analysis applications. Generally, the main contributions are:
\begin{itemize}
  \item Considering the spatio-temporal and semantic information in the trajectory data, we decouple the sequence data into several modals such as movement, route selection and global semantic characteristic and process them respectively to obtain independent embeddings which will be feed into our model.
  \item We present a representation learning model for trajectory embeddings based on multimodal fusion and attention mechanism, which combines the features from several modalities to produce an effective representation of a trajectory for the subsequent tasks.
  \item We conduct extensive experiments like unsupervised cluster task and supervised classification task on the real world trajectory data to evaluate the model. The result justifies that the ability of our model in the representation generation.
\end{itemize}

The rest of paper is organized as follows. Section~\ref{sec:review} reviews related work. Section~\ref{sec:model} proposes the DouFu framework and its details. Section~\ref{sec:expr} shows experimental evaluation on our model and section~\ref{sec:conclusion} provides the conclusion.

\section{Related Work}
\label{sec:review}
We would like to briefly introduce three major parts of related work on trajectory representation learning, multimodal learning and attention mechanism.

\subsection{Representation Learning in Trajectory}
Representation learning is the basis for intelligent computational models. A good representation can effectively reveal the internal relationships and differences between the original factors. Recently, research in representation learning has been extended to trajectory sequence data. Various models have been developed to capture the internal representation from them. Deep learning models like recurrent neural networks and one-dimensional convolutional neural networks are applied to time series data created by sliding windows\cite{dong2016characterizing, dong2017autoencoder}. Kieu et al.\cite{kieu2018distinguishing} consider trajectory data as an image which is feed to convolutional neural network to learn embeddings to represent trajectories. Ren et al.\cite{ren2020st} extract manually defined patterns from data and employ the siamese networks to them which is capable train a metric to measure the similarity between historical trajectories. This line of research is limited to the physical environment variables. None of these studies recognized the importance of functional semantic information in trajectories, which may contribute a lot to the performance of embedding generation. Actually, it is more significant to learn patterns such as longest stay point, daily commuting schedule etc. Gao et al.\cite{gao2017identifying} develop a sequence split model to group GPS location points into subsequences. Adapted from the word2vec method, this model creates check-in embeddings for each subsequence in order to learn a representation. Besides spatial patterns from location information, extensive researches focus on utilize semantic data to produce an understandable representation. Ying et al.\cite{ying2011semantic, ying2014mining} present semantic trajectory mining methods for location prediction. Most of current models prefer to incorporate spatial-temporal records with POI to extract more explicit information\cite{zhou2018trajectory, cao2019habit2vec}. But more data sources are capable of providing detailed information. In addition to POI, other mobility context can be applied to improve the quality of embeddings. Fu et al.\cite{fu2020trembr} propose a learning model, Trembr, based on road segment embeddings from geometric map matching. Trembr uses prediction of road segments cooccurence in the same trajectory to learn a representation of them. And then only feed the road segments sequence to an encoder-decoder model. Zhou et al.\cite{zhou2021self} design a contrastive mobility learning model with data augmentation for a self-supervised spatial temporal context learning

However, these methods rarely mention the route selection semantics. More features, including geometries and semantics, can be exploited in an inexpensive way than what these models do. In the context of mobility, time series information matters in semantic analysis, which could reveal the drivers’ personalized route choice paradigm. Besides statistic spatial feature, dynamic temporal data provides more flexible details and a new perspective to inspect the driving behavior and mobility pattern.

\subsection{Multimodal Learning}
Building Models with multimodal data is a significant topic that has been extensively studied in recent years\cite{baltruvsaitis2018multimodal}. Generally, a modality refers to the way to capture information about an object independently. More practically, modalities mean different information sources from a same event which can help us to know more about it. Different specific methods are applied to data captured from different modalities which will be mapped to similar feature space. And it is natural to use convolutional neural networks to process spatial visual information like videos while using recurrent neural networks to process sequential sound information like audios. The multimodal method has a variety of applications, especially in multimedia area such as human emotion classification\cite{wollmer2010context, kahou2016emonets} and video-subtitle alignment\cite{xu2015jointly,yu2016video} because of massive multi-source data.

Recently, a considerable amount of research has been presented which combine multimodal data to solve the trajectory representation learning problem. GPS, cameras, motion sensors or many other devices, like visual and auditory data source in the multimedia, collect a lot of mobile information about trajectories. It is possible to use multimodal learning methods in the trajectory analysis. Visual images and driving operation recorders can also contribute to the modality coordination and fusion\cite{PhanMinh2020CoverNetMB}. Cui et al.\cite{Cui2019MultimodalTP} apply convolutional neural networks to sky-view scene with driving states to make an online prediction about the next location of the trajectory. Chen et al.\cite{chen2020drive} employ surveillance camera data to track the vehicle in order to recover a sparse driving trajectory. But the majority of present methods involve using more sensor devices such as video cameras, which is not very convenient and inexpensive. It is easier to use existing spatial records like maps instead of sensors as a modality. DiversityGAN\cite{huang2020diversity} attempt to integrate trajectory features with map embedding vector which is generated from nearby lane coefficients. Feng et al.\cite{feng2018deepmove} propose a model called DeepMove which employs jointly multimodal embedding module with manually designed spatiotemporal and personal features to learn a dense representation. Basically, features from different modalities will be projected to similar feature spaces and then coordinated into the same one.

Nevertheless, these studies ignore the relations between modalities in the feature engineering. Models learn the weights to map modality features into several similar feature spaces and then coordinate them to build a unified feature space for subsequent usages. But it is difficult to extract knowledge from spatial temporal modal independently because of sparsity and long term dependencies. Potentially, the relationships between modalities helps to correct their biased perceptions of objects while learning the mapping weights, especially for sparse spatial temporal data.

\subsection{Attention Mechanism}
Recently, the self-attention mechanism has attracted much attention\cite{vaswani2017attention} and performs incredibly well in many learning tasks, especially in the natural language processing methods like BERT\cite{devlin2019bert} and XLNet\cite{Yang2019XLNetGA}. The attention module aims to simply calculate the correlation between the query and value to produce a weight, then applies attention weights to origin data in order to emphasize more important regions in information flow which can improve the model performance. Attention computation can resolve long term dependencies about time series data than trivial recurrent neural networks. In the field of spatialtemporal data mining area, recurrent neural network may fail to process handle the risk of gradient vanish aroused by such objects like long trajectories. Therefore, a large number of emerging methods with attention mechanism have been proposed. Gao et al.\cite{gao2019varattn} propose a generative model to extract attention of POIs from historical trajectories in order to learn representations. DeepMove\cite{feng2018deepmove} designs a historical attention module to select the most similar historical trajectory to match the current one. STDN\cite{yao2019revisiting} adopts a shifted attention mechanism to get temporal periodic information in trajectory mining.

Several studies have been conducted to use different attention information from many independent modalities. ST-LBAGAN\cite{yang2021stlbagan} uses a bidirectional attention method to learn a feature map for missing traffic data imputation. Yu et al.\cite{yu2019deep} designs a modular co-attention network which utilize the attention from video input to decode the answer from attention of given questions in order to to complete the video question answering task. In addition, LXMERT\cite{tan2019lxmert} employs a cross-modality encoder to extract mutual cross attention by encoding with each other. It is potential to analysis the multimodal trajectories data in the similar way because of extensive spatio-temporal and semantic dependencies

\section{Preliminaries}
\label{sec:preliminaries}
In this section, we first give definition for several important items and then formally formulate the trajectory representation problem.
\begin{figure*}[htbp]
  \centering
  \includegraphics[width=\textwidth]{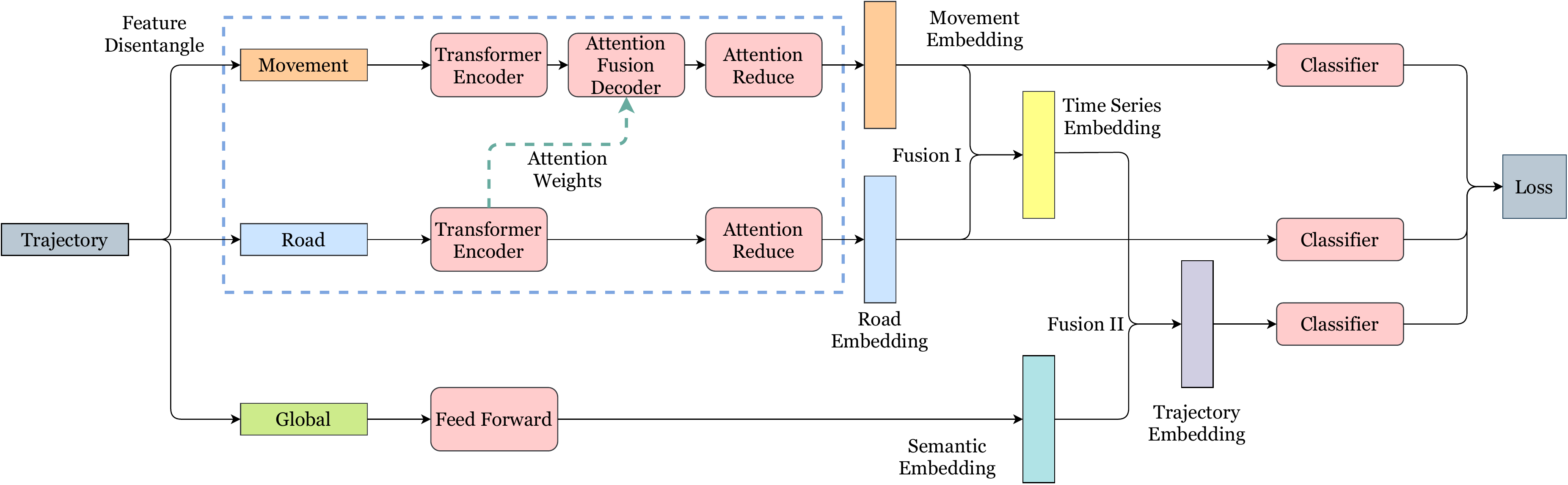}
  \caption{DouFu Framework}
  \label{fig:framework}
\end{figure*}

\begin{definition}{(Driving Trajectory).}
  The GPS device on the vechicle can record the driver's location during the travel with a sampling rate. Usually, a GPS point $p$ consists of a location point with longtitude \emph{lat} and longtitude \emph{lng} and the current time stamp $t$, i.e.$p = \{lat, lng, t\}$. A driving trajectory $\tau$ is a sequence composed of GPS points from such a GPS recorder, i.e. $\tau = \{p_1, p_2, \dots , p_n\}$.
\end{definition}

\begin{definition}{(Trajectory Set).}
  Given a user $u$ and a set of historical trajectory $\mathcal{T} = \{\tau_1, \tau_2,\dots,\tau_n\}$, a trajectory set $\mathcal{S} = \{u, \mathcal{T}\}$. A user-trajectory classification task aims to identify the true user of a new trajectory with the historical trajectory set as training data.
\end{definition}

\begin{definition}{(Road Segment).}
  In the urban traffic environment, all roads are connected to each other to construct a directed network $G = \{V, E\}$. A road segment is a edge with several attributes from the edge set $E$ which can be denoted as $r = \{l_{in}, l_{out}, a\}$ where $l_{in}$ means the in-edge set and $l_{out}$ means the out-edge set.
  And an road interaction $i$ is a vertex from the vertex set $V$.
\end{definition}

\begin{definition}{(Route).}
  Driving from origin to destination is actually along the road segments in the network while recording location with GPS. Hence a route $R$ can be defined as a sequence of road segments, i.e. $R = \{r_1, r_2, \dots, r_n\}$.
\end{definition}

\begin{definition}{(Functional Zone).}
  The travelers may want to get the service at the destination. And the proportion of functions types in the origin/destination neighborhood can represent the type of service at the current location. A functional zone vector $z_o = \{f_1, f_2, \dots, f_n\}$ where $f_i$ is the $i-$th function and $\sum_{i}^n f_i = 1$.
\end{definition}

\begin{problem}{(Driving Trajectory Representation Learning)}
Given trajectory sets $\{\mathcal{S}_1, \mathcal{S}_2,\dots, \mathcal{S}_n \}$ from users $\mathcal{U} = \{u_1, u_2, \dots, u_n\}$, train a model $M$ which is able to transform trajectories into vectors in a feature space $\mathcal{H}$. For each trajectory $\tau_u$ belonging to user $u$, we can output an embedding $\epsilon_{\tau} = M(\tau_u) \in \mathcal{H}$ where $ \mathop{Min}\limits_{i \in \mathcal{U}} \sum_{\tau \in \{\mathcal{S}_u\}} Dist(\epsilon_{\tau}, c_i) = u$.
\end{problem}

\section{The DouFu Model}
\label{sec:model}
\subsection{Framework Overview}
\label{sec:overview}
The recurrent neural network is an effective tool to analyze time series data but may be insufficient for the task of driving trajectory representation learning. There are several challenges that need to be mentioned. The first is that the driving trajectory with GPS often has a long sequence record with an average sampling rate of about 10 seconds or more per point. Under such circumstances, recurrent neural networks are less likely to learn long-term and sparse dependencies. Further, an itinerary can also be converted into modal data of various lengths respectively, like spatially located points and POI check-ins. But it is difficult for the recurrent neural networks to obtain correlations between these sequences data during processing. Inspired by multimodal learning and the attention mechanism, we propose a novel double fusion model for trajectory representation learning.

Before learning about the model, it is fundamental to recognize that a single trajectory can be regarded as encompassing multiple modalities. In our model, a trajectory can be decomposed into three independent modalities, \emph{movement}, the \emph{route} and the \emph{global} features (See Figure \ref{fig:framework}) whose detail will be explained in Section~\ref{sec:decouple}. Like other multimodal models, we handle these modalities with different methods initially. The \emph{movement} and the \emph{route} feature, as sequential inputs, contain rich time series information about the driving characteristics and route selection preference. And there is a high correlation between them because driving behavior is limited to the current road segment. Here we use a transformer encoder to extract the attention weights from them and then design an attention fusion decoder to combine these two attention weights in order to create a mixed time series embedding for the trajectory. With the \emph{global} feature which is semantically rich, a feed-forward neural network is used to generate a semantic embedding to explore the mobility pattern 245 of users. In this part, we consecutively use two methods of information fusion, including attention fusion and embedding fusion. Then, the time series and the semantic feature are merged to produce a final comprehensive embedding for the trajectory which explores the driving behavior, route selection preferences and mobility patterns. After that, we train three independent linear classifiers that identify which driver the trajectory belongs to for the \emph{movement} feature, \emph{route} feature, and final feature to learn a better representation in a supervised learning manner.

\subsection{Feature Modality Decoupling}
\label{sec:decouple}
As Section 4.1~\ref{sec:overview}, the trajectory can be independently and respectively viewed as the spatial movement sequence, the road segment sequence, and the global semantic information. We will introduce feature engineering in detail as follows.
\begin{table}[htbp]
  \centering
  \begin{tabular}{ll}
    \hline \hline
    Feature             & Variable                   \\ \hline
    Bounding Box        & Location                   \\
                        & Edge Length                \\
                        & Area                       \\
                        & Direction                  \\
    Location            & Origin                     \\
                        & Destination                \\ \hline
    Intersection Number & In                         \\
                        & Out                        \\ \hline
    Other Attribute     & Function Zone of Buffer    \\
                        & Road Segment Length/Width  \\
                        & Point Number               \\
                        & Lane Number                \\
                        & One Hot Code of Road Class \\ \hline \hline
  \end{tabular}
  \caption{Road Segment Features}
  \label{tab:route}
\end{table}

\emph{Movement}~Like the method proposed by \cite{dong2016characterizing}, it is appropriate to apply a sliding window to raw GPS data points to generate a physical movement feature sequence. We then calculate the statistics of the location points in the window as a unit, including mean, min, max, std and quantile(25\%, 50\%, 75\%) values of the speed norm, the acceleration norm, the speed difference norm, the acceleration difference norm and the angle speed norm to construct the \emph{movement} vector. From this feature, it is easy to tell the driving preference of users.

\emph{Route}~Generally, it is obvious that we can convert a trajectory into a road segment sequence by extracting each segment it pass one by one. As shown in the table~\ref{tab:route}, for each road segment, fixed length feature vector composed of road attributes which includes road start/end position, road length/direction, bounding box length/area, start/end point intersection number and other like road level and lane number. With the sequence of road segments as road feature, we are capable of analyzing the route selection preference of drivers.

\begin{table}[htbp]
  \centering
  \begin{tabular}{ll}
    \hline \hline
    Feature            & Variable                      \\ \hline
    Statistic          & Speed                         \\
                       & Difference of Speed           \\
                       & Acceleration                  \\
                       & Difference of Acceleration    \\
                       & Angle Speed                   \\ \hline
    Location           & Origin                        \\
                       & Destination                   \\ \hline
    Bounding box       & Edge Length                   \\
                       & Area                          \\ \hline
    Other Attribute    & Direction                     \\
                       & Length                        \\
                       & One Hot Code of Depature Time \\
                       & Duration                      \\
                       & Function Zone of Buffer       \\ \hline
    Road Segment(Mean) & Lane Number                   \\
                       & Road Length/Width             \\
                       & Point Number                  \\
                       & In/Out Intersection Number    \\ \hline \hline
  \end{tabular}
  \caption{Global Features}
  \label{tab:global}
\end{table}

\emph{Global}~Apparently, trajectories reveal the intention of travelers who may need the services of the destination. The description of trajectory regardless of spatial and temporal information could also represent travel patterns. In table~\ref*{tab:global}, the fix length vector, as global feature, includes departure time, O/D locations and some statistics results of move and Rote features. What is most important is the proportion of land use types\cite{gong2020landuse} in the O/D functional zone which represents the function difference between origin and destination. This can demonstrate the mobility intention of the driver in a more inexpensive way than demographic data or other sensor data.

\subsection{Road Segment Geometric Feature Pre-training}
\label{sec:pretrain}
From a driver's perspective, the route may be in the form of a road segment sequence. However, in the urban transportation environment, all roads connect with each other to build a network. Other road segments not in the driver's route could also make a difference to the current decision about route selection. Hence it is necessary to consider other segments besides those in the route when constructing the \emph{route} model. Inspired by word2vec\cite{mikolov2013word2vec} method, it may be better to create a fixed-length pre-trained feature for road segments like word vector. In order to solve this problem, we apply variational graph auto encoder(VGAE)\cite{kipf2016variational} to generate a geometric embedding instead of the hand-designed feature as the model input.

\begin{figure}[htbp]
  \centering
  \includegraphics[width=\columnwidth]{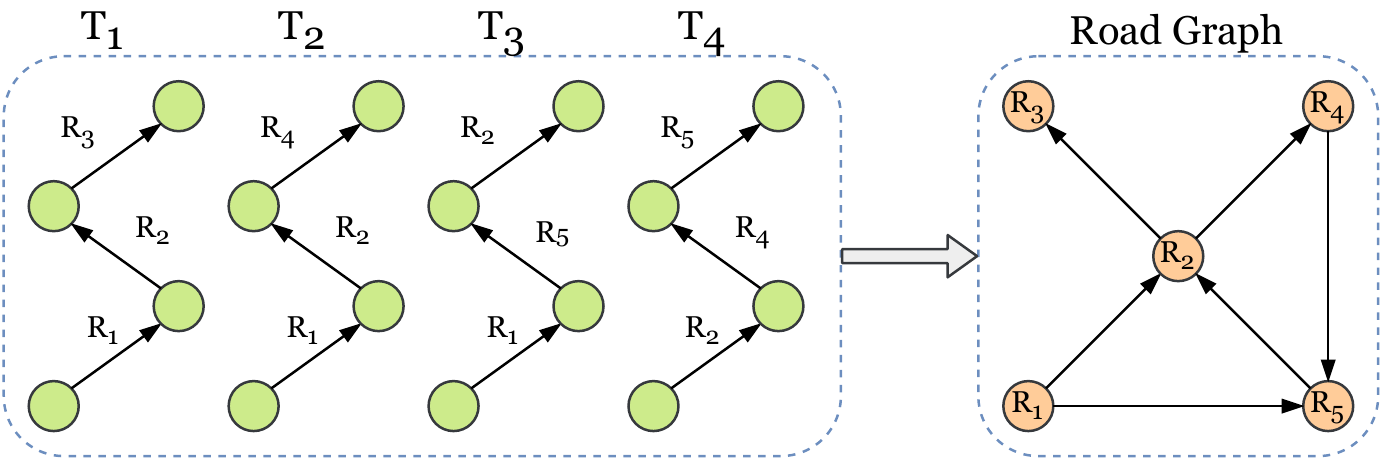}
  \caption{Generation of Road Graph}
  \label{fig:road}
\end{figure}

Firstly, we have to make a road network from segments of historical trajectories. As Figure~\ref{fig:road} shows, there is a trajectory set~$T = \{\tau_1, \tau_2, \tau_3, \tau_4\}$ and road segment set~$R = \{R_1, R_2, R_3, R_4, R_5, R_6\}$ where $\tau_1 = \{R_1 \rightarrow R_2 \rightarrow R_3\} , \tau_2 = \{R_1 \rightarrow R_2 \rightarrow R_4\}, \tau_3 = \{R_1 \rightarrow R_5 \rightarrow R_2\}, \tau_4 = \{R_2 \rightarrow R_4 \rightarrow R_5\}$. And then create a graph $G = \{V = \{R_1, R_2, R_3, R_4, R_5\}, E, X\}$ that uses road segments as vertex set $V$ and interactions as edge set $E$ with road segments features $X$.

Secondly, the embedded road segment is encoded using a variational graph auto encoder that can be trained further during the subsequent modeling process. In our method, a two-layer graph convolution network is adopted to estimate the parameters $\mu$, $\delta$ of the conditional Gaussian probability distribution $p$ for each vertex. Afterward, a latent variable $z$ is created to represent such a vertex by sampling from this distribution. For edge link from vertex $i$ to vertex $j$, a dot product probability of $z_i$, $z_j$:
\begin{equation}
  p\left(E_{i j}=1 \mid \mathbf{z}_{i}, \mathbf{z}_{j}\right)=\sigma\left(\mathbf{z}_{i}^{\top} \mathbf{z}_{j}\right)
\end{equation}
where $\delta(\cdot)$ is the logistic sigmoid function will indicate the link probability value between $i$ and $j$. And the prediction edge set:
\begin{equation}
  p(\mathbf{E} \mid \mathbf{X})=\prod_{i=1}^{N} \prod_{j=1}^{N} p\left(E_{i j} \mid \mathbf{x}_{i}, \mathbf{x}_{j}\right)
\end{equation}
which could be trained with the ground truth edge adjacency matrix later. After training, all road segments will be converted into fixed length vectors with geometric information

\subsection{Attention Fusion \& Reduction}
\begin{figure}[!t]
  \centering
  \includegraphics[width=0.4\textwidth]{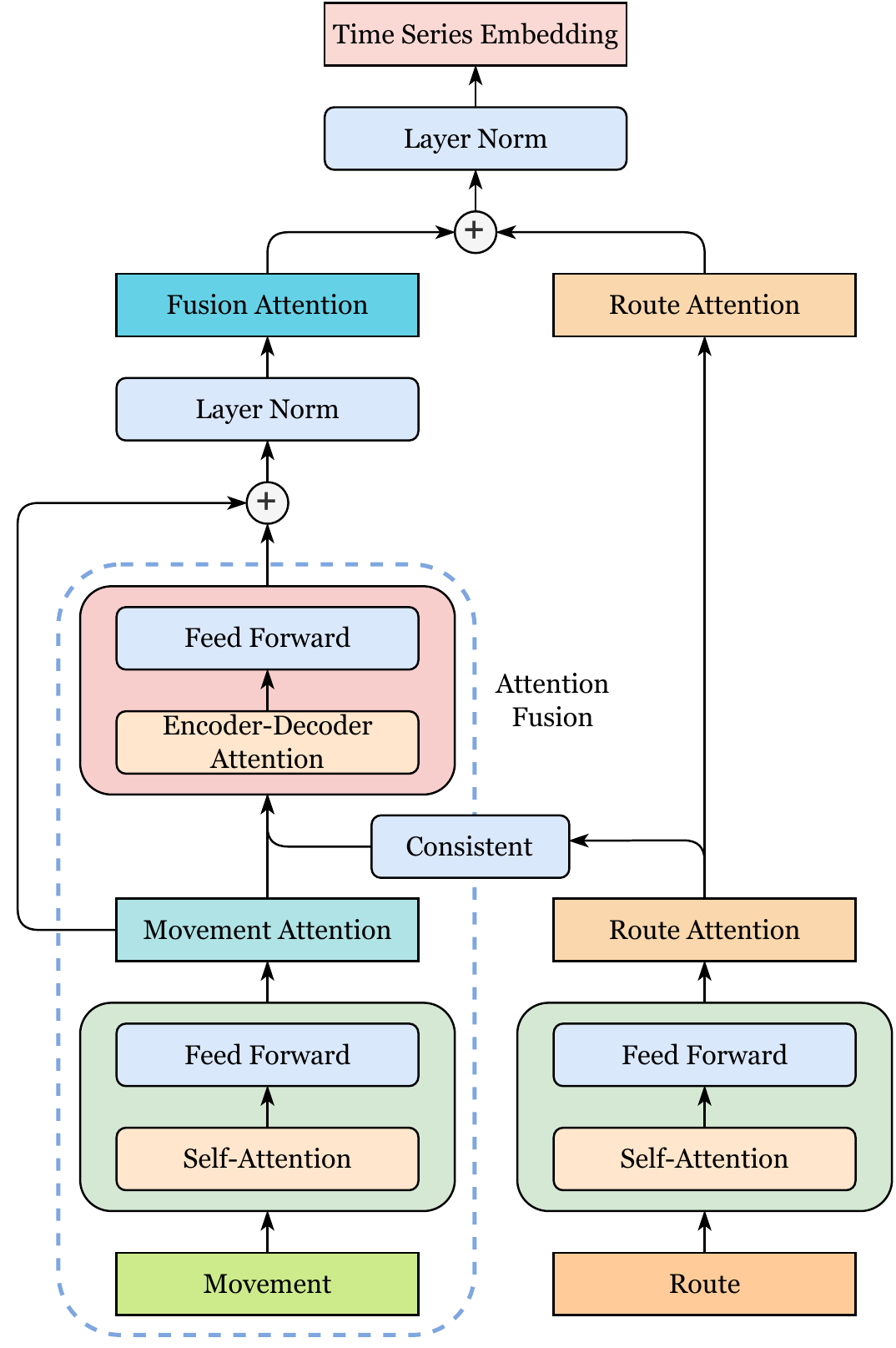}
  \caption{Road Graph}
  \label{fig:coattn}
\end{figure}

After being processed in Section~\ref{sec:decouple} and~\ref{sec:pretrain}, the movement feature and the route feature will be feed into the transformer encoder respectively to calculate the movement attention $A_m$ and route attention $A_r$ of dimension $d$. Clearly, the situation on the road is greatly influenced by the location of the vehicle. All routes are restricted by roads and can only travel along the road. Therefore, the movement attention from GPS sampling points, as a continuous signal with intensive noise, could be improved by the route attention from a deterministic finite road set. Figure~\ref{fig:coattn} shows the architecture of attention fusion decoder. Before the fusion procedure, a consistent layer is applied to map the movement feature into the shape of the route feature. Then we apply a fusion transformer decoder to decode movement attention with route attention, and learn a fusion feature by adding original movement vector:
\begin{equation}
  \text { FusionAttn }=\operatorname{LN}\left(\operatorname{SM}\left(\frac{\operatorname{C}\left(A_{m}\right) \cdot A_{r}}{\sqrt{d}}\right) A_{m}+A_{m}\right)
\end{equation}
Where $\operatorname{LN}$ is a layer norm operator, $\operatorname{SM}$ is a softmax operator, and $\operatorname{C}$. is a consistent layer. After fusion, the sequence length of movement feature will be the same as the one of route. Inspired by attention reduce\cite{yu2019deep}, multiple linear layers are adopted to learn the weight of each feature to produce a fixed length embedding from fusion features:
\begin{equation}
  \alpha_{i}=\operatorname{softmax}\left(\operatorname{MLP}_{\mathrm{i}}(X)\right)
\end{equation}
\begin{equation}
  \tilde{x}=\sum_{i=1}^{n} \operatorname{Layer} \operatorname{Norm}\left(\sum_{j=1}^{m} \alpha_{i j} x_{j}\right)
\end{equation}

Where $\alpha_i$ is the $i$-th weight for feature $X = \{x_1, x_2, \dots, x_m\}$. And the weighted sum of features can be obtained as the representation result. We use cross entropy as the loss function to train three classifiers to identify the drivers of the current trajectory on the top of the comprehensive feature:
\begin{equation}
  \mathcal{L}=\mathcal{L}_{\text {fusion }}+\alpha \mathcal{L}_{\text {move }}+\beta \mathcal{L}_{\text {route }}
\end{equation}
Where $\alpha$ and $\beta$ are the learning rate adjust factors.

\section{Experimental Evaluation}
\label{sec:expr}

\subsection{Experimental Details}

\begin{figure*}[htbp]
  \centering
  \begin{minipage}[t]{0.45\textwidth}
    \centering
    \includegraphics[width=0.75\textwidth]{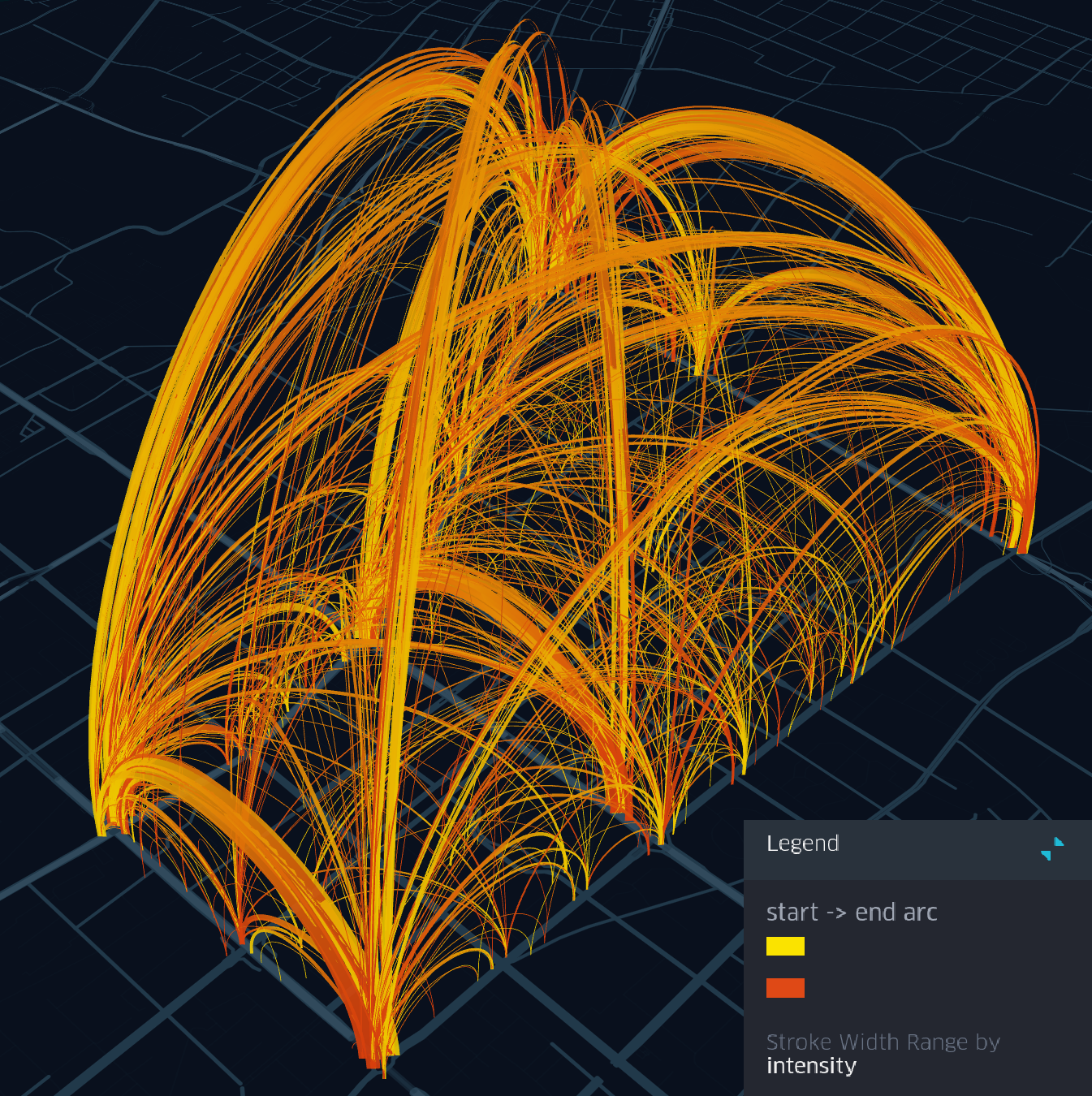}
    \caption{Evaluation Data Area}
    \label{fig:data}
  \end{minipage}
  \begin{minipage}[t]{0.49\textwidth}
    \centering
    \includegraphics[width=1\textwidth]{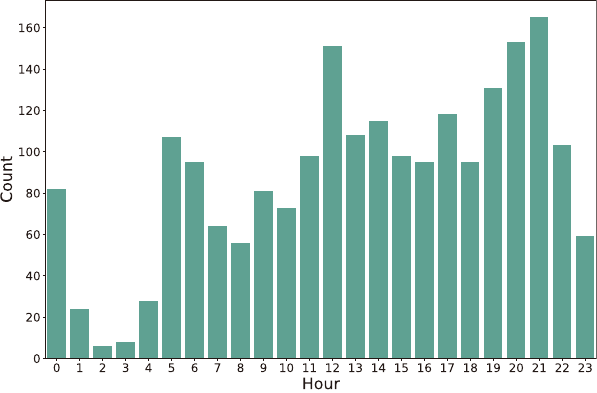}
    \caption{Departure Time Distribution}
    \label{fig:time}
  \end{minipage}
\end{figure*}

In this section, a private dataset is adopted from a navigation service company to train and evaluate our model with several baselines. This dataset collects a large number of driving trajectories in a subarea of Beijing in December 2018 as Figure~\ref{fig:data} shows. The departure times of these trajectory data are distributed over a 24-hour period in Figure~\ref{fig:time}. The trajectories were firstly divided into training sets and test set according to users. We selected 100 drivers with 10,575 trajectories as the training set. Any trajectory of users in the training set will not be in the test set. Then we selected 21 drivers with 2,113 trajectories as testing set. The training set will be split by 8:2 for training and 315 validation. From these trajectories data, we created a road segment graph $G$ with 4,158 vertices and 33,731 edges for geometric learning.

To demonstrate the performance of DouFu which has been trained, we use embeddings with the same size from the test set to conduct the classification and the clustering task. In the classification task, some simple machine learning 320 classifiers like linear regression and SVM are adopted to finish the trajectoryuser classification with supervised training. In the clustering task, unsupervised methods like K-means are applied to group all trajectory representations into clusters. And we will evaluate both of the experiment results

\subsection{Evaluation Metrics \& Baselines}
\subsubsection*{Evaluation Metrics}
\textbf{Classification}~This task intends to predict the candidate users for the input trajectory. We use accuracy and macro-F1 to measure the performance, which are common metrics for classification tasks.
\begin{itemize}
  \item \textbf{Accuracy}:\begin{equation*}
          ACC = \frac{\#\text{true prediction of trajectories}}{\#\text{trajectories}}
        \end{equation*}
  \item \textbf{macro-F1}:\begin{equation*}
          \text{macro-F1} = \frac{2 * P * R}{P + R}
        \end{equation*}
        where $P$ is average precision and $R$ is recall value.
\end{itemize}

\textbf{Clustering}~The clustering task tries to divide trajectories into groups. We apply Davies-Bouldin index, normalized mutual information score and adjusted rand score to compare the clustering result with the original trajectory sets $\mathcal{S}$.
\begin{itemize}
  \item \textbf{Davies-Bouldin Index} alculate the ratio of within-cluster distances to between-cluster distances in order to measure the similarity of clusters which means a better clustering result is supposed to have a lower index value:\begin{equation}
          \mathrm{DB}=\frac{1}{n} \sum_{i=1}^{n} \max _{j \neq i}\left(\frac{\sigma_{i}+\sigma_{j}}{d\left(c_{i}, c_{j}\right)}\right)
        \end{equation}
        where $c_i$ is the center of $i$-th class, $\delta_i$ is the mean distance between sample in the $i$-th class and $c_i$, and $d(c_i, c_j)$ is the distance between $c_i$ and $c_j$
  \item \textbf{Normalized Mutual Information Score} can estimate the correlation between two cluster results:\begin{equation}
          \operatorname{NMI}(\Omega, C)=\frac{I(\Omega ; C)}{(H(\Omega)+H(C) / 2)}
        \end{equation}
        where $I$ is the mutual information and $H$ is the entropy. Similar objects $\Omega$ and $C$ will result in a high NMI value.
  \item \textbf{Adjusted Rand Score} regards cluster as a decision process:\begin{equation}
          \mathrm{ARI}=\frac{\mathrm{RI}-E[\mathrm{RI}]}{\max (\mathrm{RI})-E[\mathrm{RI}]}
        \end{equation}
        where \begin{equation}
          R I=\frac{T P+T N}{T P+F P+T F+F N}=\frac{T P+T N}{C_{N}^{2}}
        \end{equation}
        A higher rand score indicates that the clustering results may be consistent with the ground truth.
\end{itemize}

\begin{table*}[htbp]
  \centering
  \caption{User Prediction Performance}
  \label{tab:classification}
  \resizebox{\textwidth}{!}{%
    \begin{tabular}{ccccccccccccccc}
      \hline
      \multirow{2}{*}{Model}                &
      \multicolumn{2}{c}{Linear Regression} &
      \multicolumn{2}{c}{Ridge Regression}  &
      \multicolumn{2}{c}{Navie Bayes}       &
      \multicolumn{2}{c}{SVM}               &
      \multicolumn{2}{c}{KNN}               &
      \multicolumn{2}{c}{MLP}               &
      \multicolumn{2}{c}{Decision Tree}                                                                                                                                                                                                                                  \\
                                            & ACC            & F1             & ACC            & F1             & ACC             & F1             & ACC   & F1    & ACC            & F1             & ACC            & F1             & ACC            & F1             \\ \hline
      RNN Move                              & 28.67          & 25.93          & 27.99          & 22.34          & 27.31           & 23.79          & 20.15 & 17.73 & 26.37          & 24.83          & 29.35          & 28.03          & 22.45          & 21.96          \\
      RNN Route                             & 17.78          & 14.07          & 17.38          & 11.83          & 16.84           & 9.97           & 9.46  & 6.78  & 13.39          & 11.79          & 16.84          & 14.53          & 12.64          & 11.39          \\
      Global                                & 12.70          & 11.84          & 10.95          & 6.94           & 9.73            & 7.02           & 7.91  & 5.71  & 6.82           & 5.32           & 13.92          & 11.84          & 7.43           & 6.65           \\ \hline
      RNN Fusion                            & 31.68          & 29.38          & 28.91          & 23.57          & 25.39           & 22.78          & 21.19 & 18.91 & 28.64          & 26.61          & 30.40          & 28.80          & 22.35          & 22.00          \\
      Semantic Fusion                       & 40.64          & 39.05          & 37.73          & 32.03          & 33.61           & 30.51          & 30.90 & 31.15 & 30.56          & 28.40          & 35.23          & 34.43          & 21.43          & 21.64          \\
      Attention Fusion                      & 42.26          & 41.20          & 37.19          & 31.57          & 37.46           & 34.25          & 33.06 & 32.42 & 34.01          & 31.79          & 37.39          & 36.40          & 24.38          & 24.42          \\
      \textbf{Double Fusion}                & \textbf{43.81} & \textbf{42.03} & \textbf{41.45} & \textbf{35.33} & \textbf{ 39.01} & \textbf{35.85} & 31.10 & 29.47 & \textbf{34.41} & \textbf{32.34} & \textbf{40.56} & \textbf{36.85} & \textbf{24.47} & \textbf{24.61} \\ \hline
    \end{tabular}%
  }
\end{table*}

\textbf{Baseline Algorithms}
It is not appropriate to compare our model which is in an multimodal manner with other models that use a single input like DeepMove\cite{yu2019deep}. Hence we design several baseline algorithms in order to demonstrate the performance of modules in our method.
\begin{itemize}
  \item \textbf{RNN Move}\cite{dong2016characterizing} applies recurrent neural network to the movement feature to generate a embedding.
  \item \textbf{RNN route}, which borrows the idea from RNN Move, feeds the route feature to the recurrent neural network to produce a time series feature about route selection.
  \item \textbf{Global} employs feed forward neural network to capture internal characteristics from global feature designed in Section~\ref{sec:decouple}.
  \item \textbf{RNN Fusion} simply feeds the movement and route sequence feature to an independent recurrent neural network and combine them with a feed forward layer.
  \item \textbf{Semantic Fusion} is a recurrent neural network based model which can be considered as a simplified version of our method. It is necessary to compare the performance of RNN and the attention module.
  \item \textbf{Attention Fusion} only applies attention mechanism and attention reduce module to process time series data and then only uses linear layer to integrate them.
\end{itemize}

\begin{figure*}[bp]
  \centering
  \includegraphics[width=1\textwidth]{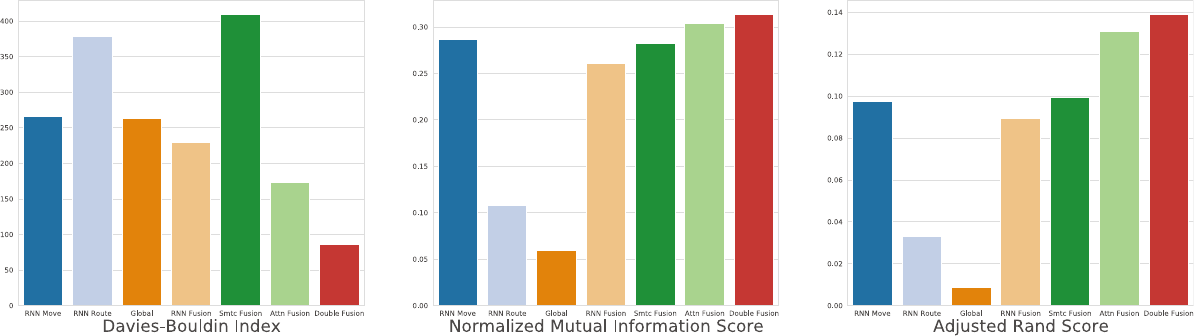}
  \caption{Clustering Performance}
  \label{fig:clustering}
\end{figure*}

\subsection{Result Analysis}

\subsubsection*{Classification}
We first train Linear Regression, Ridge Regression, Naive Bayes, SVM, KNN, MLP, and Decision Tree with embeddings from models to be evaluated in a 10-fold cross validation manner in order to complete the trajectory-user linking task. And then Table~\ref{tab:classification} evaluates them with ACC and macro-F1.

In all learners, differentiable models like MLP, regressors work better than non-differentiable ones like decision trees with the embeddings. Because the differentiable models are more capable of handling continuous and uniform signal inputs like well-trained computational representations rather than non-differentiable categorical values. Further, decision trees can detect the importance of input fields but may fail if the fields are equally important. Hence Table~\ref{tab:classification} indicates that our method learns a more uniform feature space.

The \emph{movement} and \emph{route} feature obtain a higher score while the \emph{global} feature contribute less to the representation learning which means a general pattern of trajectories may be insufficient to describe spatial-temporal streaming data. However, it does not mean semantic information is good for nothing. Compare to a simple RNN fusion model, the semantic fusion model outperforms considerably. Mobility routines and intentions from the functional zone differences matter in trajectory pattern mining. Additionally, the attention module works better than recurrent neural networks. Among most of the machine learning approaches, the Double Fusion models archive better performance than others.

\begin{figure*}[ht]
  \centering
  \includegraphics[width=1\textwidth]{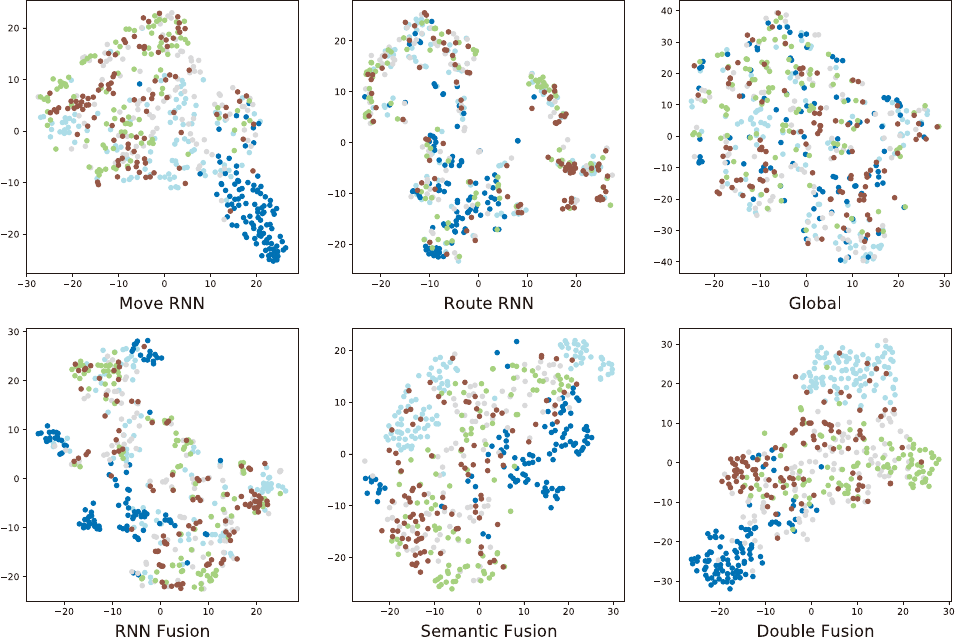}
  \caption{Embedding TSNE}
  \label{fig:tsne}
\end{figure*}

\subsubsection*{Clustering}

We utilize embeddings from such models for K-means clustering directly in the way of unsupervised learning. And then evaluate them with metrics. Figure~\ref{fig:clustering} shows the result of clustering evaluation. A lower Davies-Bouldin index value mainly indicates a better within-class clustering distribution where all embedding elements of the same class in the feature space tend to be more closely around their center. The DouFu outperforms other models, which discovers more internal associations between trajectories of the same driver than others. A better normalized mutual information score and adjusted rand score indicate a better partitioning result for trajectories of different drivers. The clustering evaluation results show that our method can provide a effective representation which is appropriate for driver's similarity and pattern analysis

Classification and clustering evaluation results demonstrate that the joint learning method with multiple inputs mentioned above can improve the quality of embeddings. Inputs from different modalities can complete the task respectively, but they tend to focus only on one aspect of data from a moving vehicle. It is obvious that a comprehensive fusion model is capable of capturing more information. In addition, the attention module can calculate correlations between independent modules, which enhances DouFu's performance.

Figure~\ref{fig:tsne} shows the embedding TSNE result of models with 4 selected users. The embeddings from DouFu model shows a better understanding about the features of trajectories than others.

\section{Conclusion}
\label{sec:conclusion}
In this paper, we propose a joint representation learning method, DouFu, for driving trajectories. The DouFu applies multimodal learning and attention fusion to the trajectory and other data in order to generate a comprehensive embeddings which is capable of capture the internal characteristics. Empirically, we evaluate models and the DouFu show a better performance than other on the classification and clustering tasks. The results show that our approach can provide effective representations for trajectories in the absence of demographic or other sensor data, using only trajectory and basic map information, which prove that route selection preferences and mobility intentions matter in the trajectory patter mining and user analysis.

For future work, there are some issues that need to be addressed. First, our model encodes spatial location in the terms of relative position of bounding box without considering position correlation. We plan to utilize special representation learning methods for spatial location of objects. Second, while processing time series feature, a simple transformer encoder and decoder are applied. A cross-modal transformer or more complex model can be adopted to improve the performance of our model. Third, as a multimodal model, it is possible to exploit more global spatial temporal data to produce a more detailed representation. We can use traffic data to generate time slices of road segment features.

\section{Acknowledgements}
This research was supported by grants from the National Key Research and Development Program of China (2017YFE0196100), and the National Natural Science Foundation of China (41771425, 41830645, 41625003). We also appreciate the detailed comments from the Editor and the anonymous reviewers.

\bibliography{mybibfile}

\end{document}